\documentclass[runningheads]{llncs}
\usepackage[T1]{fontenc}
\usepackage{placeins} 
\usepackage{float}
\usepackage{graphicx}
\usepackage{tikz}
\usepackage{soul}
\usepackage{wrapfig}
\usepackage[misc,geometry]{ifsym}
\usepackage{subfigure}
\usepackage{amsfonts}
\usepackage{xcolor}
\usepackage[nolist,nohyperlinks]{acronym}
\acrodef{LLM}{Large Language Model}
\acrodef{VLM}{Vision Language Model}
\acrodef{VLMs}{Vision Language Models}
\acrodef{LLMs}{Large Language Models}
\begin{document}
\title{Task-oriented Robotic Manipulation with Vision Language Models\thanks{This work is supported by the EPSRC National Edge AI Hub (EP/Y007697/1).}}
%
%
\author{Nurhan Bulus Guran\inst{1} \and
Hanchi Ren\inst{1}\and
Jingjing Deng\inst{2}\and
Xianghua Xie\inst{1}\textsuperscript{\Letter}}
\authorrunning{N.B.Guran et al.}
%
\institute{Department of Computer Science,
        Swansea University, Swansea, United Kingdom. \\ 
\and
Department of Computer Science, 
        Durham University, Durham, United Kingdom.\\
\email{x.xie@swansea.ac.uk}  
}
\maketitle              
\thispagestyle{empty}
\pagestyle{empty}
\begin{abstract}
\ac{VLMs} play a crucial role in robotic manipulation by enabling robots to understand and interpret the visual properties of objects and their surroundings, allowing them to perform manipulation based on this multimodal understanding. Accurately understanding spatial relationships remains a non-trivial challenge, yet it is essential for effective robotic manipulation. In this work, we introduce a novel framework that integrates \ac{VLMs} with a structured spatial reasoning pipeline to perform object manipulation based on high-level, task-oriented input. Our approach is the transformation of visual scenes into tree-structured representations that encode the spatial relations. These trees are subsequently processed by a \ac{LLM} to infer restructured configurations that determine how these objects should be organised for a given high-level task. To support our framework, we also present a new dataset containing manually annotated captions that describe spatial relations among objects, along with object-level attribute annotations such as fragility, mass, material, and transparency. We demonstrate that our method not only improves the comprehension of spatial relationships among objects in the visual environment but also enables robots to interact with these objects more effectively. As a result, this approach significantly enhances spatial reasoning in robotic manipulation tasks. To our knowledge, this is the first method of its kind in the literature, offering a novel solution that allows robots to more efficiently organize and utilize objects in their surroundings.


\end{abstract}
\section{Introduction}
Robotic manipulation using \ac{VLMs} represents a new powerful paradigm\cite{duan2024manipulate}. The robotic manipulation term describes a robot's capacity to physically interact with items in its surroundings in order to carry out operations tasks in daily life like moving or rearranging\cite{billard2019trends}.

Understanding spatial relationships between objects and their surroundings is vital for robots in many fields, ranging from domestic assistance to industrial automation, as it enables them to function independently in dynamic, real-world environments. \ac{VLMs}, like MiniGPT-4 \cite{zhu2023minigpt} and LLaVA \cite{liu2024visual}, have emerged as a promising method for improving robotic manipulation by enabling robots to comprehend and reason about the objects they interact with. The ability of \ac{VLMs} to interpret objects' properties and to understand spatial relationships is essential for effective manipulation, as it allows robots to better perceive and interact with their environment. However, current \ac{VLMs} fall short in fully comprehending spatial relationships, which are fundamental for successful robotic manipulation tasks.

Recent advancements in robotic manipulation have shown promising results in improving the robot’s ability to recognize and interact with objects \cite{zheng2024survey}. While these methods have shown promising results, they suffer from significant limitations. \ac{VLMs} can describe objects and their surroundings, but their ability to understand spatial relationships between objects remains limited. Additionally, existing datasets often lack the comprehensive annotations needed to capture object properties such as fragility, mass, material, and transparency, which are essential for reasoning about how objects should be manipulated \cite{Kruzliak_2024}. This gap hinders the development of robotic systems capable of sophisticated spatial reasoning, such as determining the correct organization or arrangement of objects in cluttered environments.

The limitations we observed in existing approaches relate to both the datasets and the techniques employed in the current research. 
The inability of robots to comprehend intricate relationships between objects results from insufficiently developed spatial reasoning abilities in \ac{VLMs} and the lack of understanding of crucial object attributes in datasets. Furthermore, whole-scene descriptions, where object relationships are not represented, are often insufficient to capture the nuances of spatial dynamics, limiting the overall effectiveness of \ac{VLMs} in robotic tasks.

In order to address these issues, the following question will be posed in this research: 
\textit{Can the representation of spatial relationships between objects, together with their corresponding properties, be transformed into structures that enhance a robot’s ability to organize and manipulate objects in real-world environments?}

To overcome these difficulties, we introduce a new dataset that is intended to enhance spatial reasoning. About 600 images were generated, each showing different objects organized on a desk. After each object is detected, important physical attributes including mass, fragility, material, and transparency are determined using an optimized \ac{VLM}\cite{gao2024physically}. When determining the proper way for the manipulation of objects, these qualities offer vital information. In addition to detecting the objects and determining the attributes, we also manually annotated the images in order to depict the spatial relationships among the objects. 
Our main focus is on identifying the positional relationships that define their relative positions.In our proposed method, a tree structure is automatically created based on these captions, with nodes representing the objects and edges indicating their spatial relationships. These tree structures are then fed into a language model together with the properties of the item to direct the reorganization of the objects into a more structured state according to a specified task.\\
\\

Our main contributions are as follows:
\begin{itemize}

\item[\textbullet] We introduce a new dataset of 600 images with manually annotated object attributes and spatial relationships to improve spatial reasoning in robotic tasks. 

\item[\textbullet] We propose a new method that transforms hierarchical structures representing object relationships into new hierarchical structures according to the given high-level task description.

\item[\textbullet] Enhanced spatial reasoning in robotic manipulation, particularly improving the understanding and handling of complex spatial relationships.
\end{itemize}


\section{Related Work}

The ability to reason about spatial relationships is crucial for robots to effectively interact with and manipulate objects in their environment. While recent advancements in Language Models have shown promising capabilities, several limitations remain.
\subsection{Spatial Reasoning in Language Models}
SpatialVLM \cite{chen2024spatialvlmendowingvisionlanguagemodels} focuses on endowing \ac{VLMs} with explicit spatial reasoning capabilities, allowing them to answer queries about spatial arrangements in scenes. Similarly, SpatialRGPT \cite{cheng2024spatialrgptgroundedspatialreasoning} explores grounded spatial reasoning in \ac{VLMs}, enabling them to interpret spatial relationships from visual input, which is essential for scene understanding. Liao et al.\cite{liao2024reasoningpathsreferenceobjects} delve into quantitative spatial reasoning in \ac{VLMs}, examining their ability to reason about object sizes and distances, demonstrating the potential for more precise spatial understanding. These works collectively demonstrate the growing ability of \ac{LLMs} to process and reason about spatial information, forming a foundation for integrating language-based spatial understanding into robotic systems. However current approaches often struggle with explicitly representing spatial relationships in a structured manner. They lack the explicit scene representation needed for planning how to change the arrangement of objects.
\subsection{Language Models for Robotics}
The application of \ac{LLMs}  to robotics has shown significant promise in enabling robots to perform complex tasks from natural language instructions. Driess et al. \cite{driess2023palmeembodiedmultimodallanguage} introduce PaLM-E, an embodied multimodal language model that integrates language models with visual and robotic control, demonstrating capabilities in mobile manipulation and task and motion planning. RT-2 \cite{brohan2023rt2visionlanguageactionmodelstransfer} further showcases the power of vision-language-action models for robotic control, allowing robots to leverage web knowledge to perform a variety of manipulation tasks based on language commands. Wen et al. \cite{wen2024objectcentricinstructionaugmentationrobotic} explore object-centric instruction augmentation, using multimodal LLMs to enrich language instructions with object location information, improving robotic manipulation performance. These studies highlight the potential of LLMs to bridge the gap for robot actions, enabling more intuitive and flexible robotic control. They produce robot actions from multimodal inputs. Instead of directly generating actions, our approach leverages LLMs to reason about how the spatial relationships need to change to satisfy a given prompt. This provides a crucial intermediate step for action generation and allows for more nuanced and effective robotic manipulation compared to methods that directly map language to actions or focus solely on spatial description or question answering.

\section{Problem Statement}
In robotic manipulation tasks, the ability to accurately understand and interpret the spatial relationships between objects is essential for effective operation, particularly in tasks such as organizing or assembling objects in specific configurations.
While \ac{VLMs} have shown promise in integrating visual and linguistic information to understand scenes, they often fail to capture complex spatial relationships. The current state-of-the-art \ac{VLMs} rely heavily on pre-trained models that are not specifically designed to address the nuances of object interaction and spatial reasoning required for robotic tasks.

Furthermore, existing datasets and models tend to overlook the physical properties of objects, which are crucial in determining how robots should interact with them \cite{jiang2023roboticperceptiontransparentobjects}. This limitation hinders the ability of robots to perform tasks that require precise manipulation based on object properties and spatial context. To advance robotic manipulation skills, it is imperative to bridge the gap between scene interpretation and action-oriented spatial reasoning.

\begin{figure*}[ht]
\vspace{-1cm}
\begin{center}
\includegraphics[width=1\linewidth]{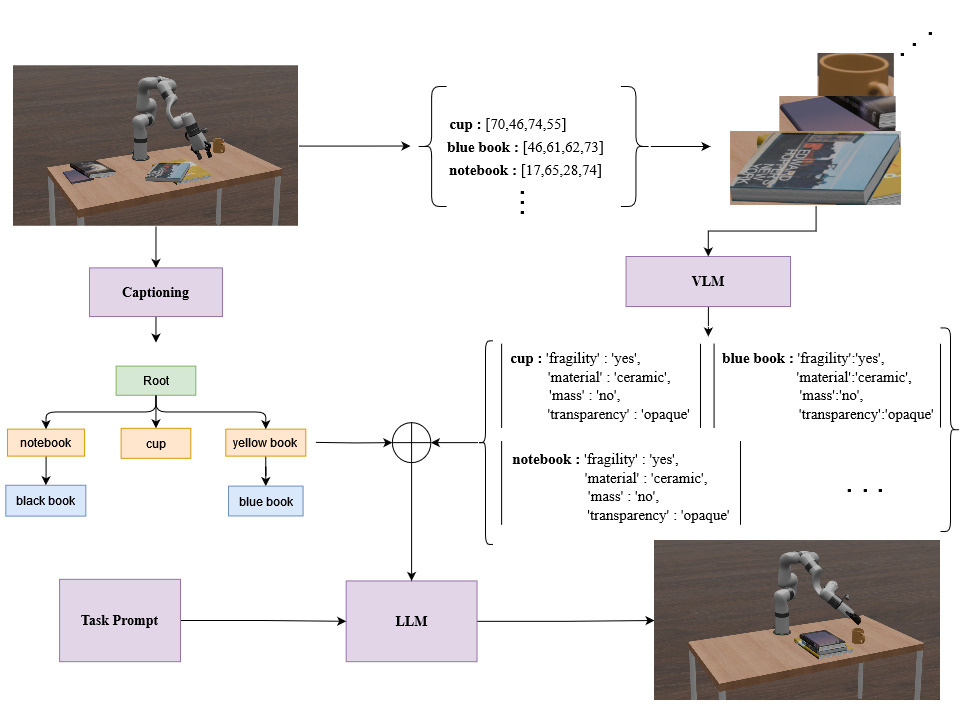}
\caption{Overview of our framework: Objects are first detected and their attributes are extracted using a fine-tuned \ac{VLM}\cite{gao2024physically}. Spatial relationships between objects are manually described and used to build tree structures representing these relationships. These tree structures are first combined with object attributes. Then this combination with a task-oriented prompt fed into a language model to generate a new representation according to the given task.}
\label{flow}
\end{center}
\end{figure*}
In order to overcome this, our research suggests developing a unique dataset and approach that are intended to improve \ac{VLMs}' spatial reasoning capabilities, particularly for robotic applications. Our goal is to increase the precision and effectiveness of robotic manipulation in dynamic, task-driven environments by integrating important object properties and hierarchical structures that depict spatial relationships. By addressing the issue of insufficient spatial thinking in existing \ac{VLMs}, this work aims to pave the way for the development of more efficient robotic systems that can perform intricate manipulation tasks.

\section{Methodology}
In this section, we present the details of the proposed methodology for enhancing spatial reasoning capabilities in robotic manipulation tasks. To improve robotic manipulation in task-specific contexts, we propose a comprehensive framework that integrates object detection, attribute assignment, and hierarchical spatial relationship modeling. The overall approach is illustrated in Figure~\ref{flow}.


\subsection{Dataset}

A dataset comprising images was generated, each depicting various objects arranged differently on a tabletop surface. To create this dataset, we utilized Blender \cite{blender2023}, an open source 3D modeling and rendering software, to simulate and render synthetic scenes featuring diverse object configurations. These images served as the foundation for analyzing the spatial relationships among the objects. An object detection model\cite{yolov8_ultralytics} was employed to identify all objects within each image.

Following detection, and inspired by the work of Gao et al. \cite{gao2024physically}, each detected object was processed using a fine-tuned \ac{VLM} to extract its key physical attributes. Specifically, the model identified properties such as fragility, representing the degree to which an object is delicate and requires careful handling; mass, referring to the object's weight, which influences the force needed during manipulation; transparency, indicating the level of opacity or clarity, which affects how the object interacts with light; and material, describing the substance from which the object is made, playing a significant role in determining its behavior under various physical conditions. These attributes are essential for informing manipulation strategies in task-oriented scenarios, enabling the robotic system to adapt its actions according to the specific characteristics of each object.

\subsection{Task-Based Reorganization and Model Application}

After detecting the objects and assigning their attributes, spatial relationship captions were manually written for each image. These captions described the positioning of each object relative to the others. The relationships between the detected objects are captured through captions associated with the images. We utilized a natural language processing (NLP) pipeline to extract these relations. This process involves analyzing the 
captions provided for each image, where objects and their spatial arrangements are described in natural language. 
The sentences were parsed and the spatial descriptions were converted into triplets that encode spatial relations in a subject–predicate–object format.


These triplets were then used to construct a hierarchical tree structure, where each object corresponds to a node containing its attributes, and directed edges represent the relationships between nodes.The process of constructing the tree followed a systematic approach. 
For example, if an "on top of" relationship was found between two objects, a directed edge was drawn from the object at the bottom (e.g., a table) to the object placed on top of it (e.g., a book), reflecting the physical arrangement between them. This approach results in a hierarchical tree structure, where the object at the lowest level in the scene, such as a table, serves as the root node, while the objects situated above it are connected as dependent child nodes. Each edge represents a meaningful spatial relationship, ensuring the hierarchy accurately mirrors the real-world arrangement of objects. The structure is designed with rules to avoid cycles and maintain logical connections, ensuring that no object can be both a parent and a child in the same loop, preserving the tree's integrity and preventing circular relationships. This tree structure provided a detailed representation of the spatial arrangement and physical properties of the objects.

The hierarchical tree structure, representing the spatial arrangements and physical properties of objects was provided as text input to the GPT-4o language model\cite{openai2023gpt4}.
The model was prompted with a specific tabletop manipulation task (e.g., “stack the book”), and its objective was to reorganize the initial tree structure accordingly, producing a more logically structured arrangement that fulfils the given task using only the provided object attributes and spatial information. Without direct visual input, the model relied solely on the hierarchical tree structure. 
Lacking any visual information about the environment, the model used only the provided hierarchical tree structure along with the attributes of the objects to perform the reorganization. The prompt given to the model was based on a scenario where a robotic arm, unable to perceive its environment visually, uses the provided tree structure to understand object relationships and perform a task. 
The prompt specified that the model could only use the given objects' labels and their respective properties to make logical decisions regarding their organization. The output expected from the model was a transformed tree structure representing the final state of the environment, with the same number of objects, ensuring each object was used only once. This process allowed the model to more efficient and coherent arrangement of the objects, resulting in a new hierarchical tree structure that better reflected the desired outcome.

This task-driven reorganization process utilized the initial and transformed hierarchical tree structures generated by the language model to derive a structured manipulation plan, simulating the rearrangement strategies a robot might employ in real-world scenarios.

\section{Evaluation}
\begin{figure}[t]
\vspace{-1cm}
\centering
    \includegraphics[width=1\linewidth]{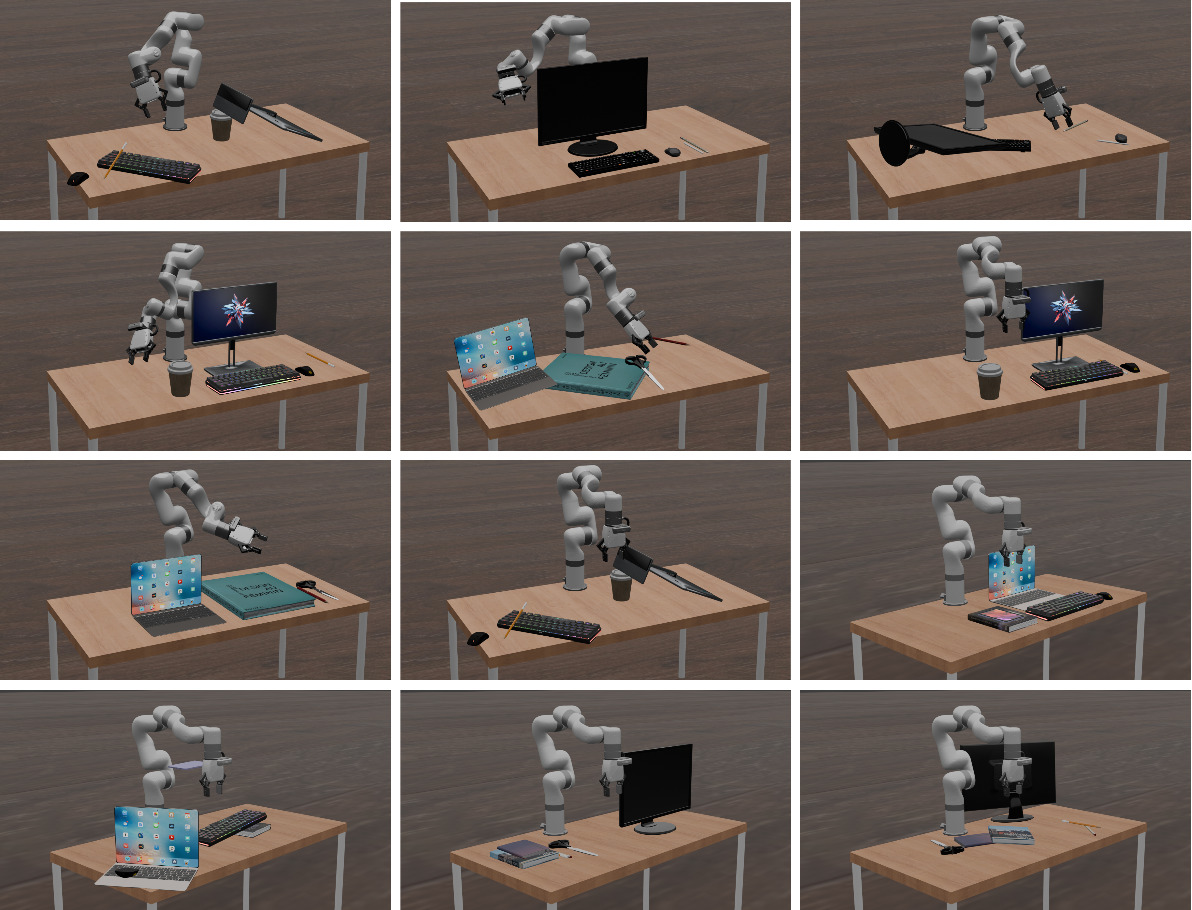}
\caption{Sample synthetic images randomly arranged objects for task-oriented robotic manipulation from the dataset.}
\label{fig2}
\end{figure}

To assess the effectiveness of our approach, we evaluate its performance using the proposed dataset, which is specifically designed to enhance spatial reasoning in robotic manipulation tasks. The dataset consists of synthetic images depicting various tabletop layouts with randomly arranged objects. These objects exhibit physical characteristics commonly found in real-world scenarios and are intended to improve the robot’s understanding of spatial relationships and task-oriented manipulation. Sample images from the dataset are shown in Figure~\ref{fig2}.

\begin{figure*}[ht]
\vspace{0cm}
\centering
\includegraphics[width=1\linewidth]{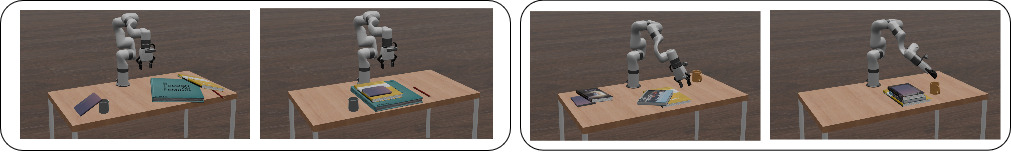}
\caption{Task Simulation Examples. Initial images represent the generated images, while the second set illustrates the simulation results.}
\label{fig3a}
\end{figure*}
\subsection{Object Understanding and Spatial Annotation}

To prepare the visual data for spatial reasoning, we first employed the object detection \cite{yolov8_ultralytics} to identify and classify each object in the image. All objects in the images were successfully detected and assigned a unique identifier for further analysis. 

Following detection, \ac{VLM} \cite{gao2024physically} was used to extract key physical attributes of each object. These include material, fragility, mass, and transparency—properties essential for informed manipulation in task-oriented scenarios. For instance, a fragile object would require careful handling, while a transparent object might influence the robot’s visual perception and decision-making\cite{jiang2023roboticperceptiontransparentobjects}. By incorporating both visual recognition and attribute extraction in a unified pipeline, this stage provides a comprehensive representation of each object, forming a foundational step for the spatial reasoning and task execution components of our framework. This representation allows the robot to make context-aware decisions, directly contributing to its ability to interact meaningfully with its environment.

\begin{figure*}[ht!]
\centering
\includegraphics[width=0.6\linewidth, height=1.0\textheight, keepaspectratio]{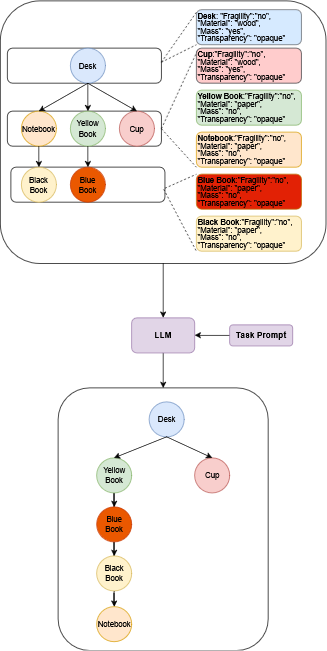}
\caption{An example of initial and transformed hierarchical tree structures.}
\label{fig3b}
\FloatBarrier
\end{figure*}

Once the individual objects and their attributes were extracted, spatial relationships between each object pair were manually annotated. The scope of relationships was intentionally limited to "on" and "on top of" to ensure clarity and consistency throughout the dataset. These relations provide foundational information for understanding object dependencies and spatial hierarchies. To capture this structure, human-generated captions were created to describe the spatial configuration in each image. These captions, along with the retrieved object attributes, were used to construct hierarchical tree structures that explicitly encode the spatial dependencies among objects. These trees serve as an interpretable and actionable guide for task-driven robotic manipulation.

\subsection{Hierarchical Tree Structure}
In constructing the hierarchical tree structure, a key priority was to develop a representation that is not only straightforward and comprehensible but also sufficiently detailed to provide meaningful input for language models. To achieve this, specific rules were defined to govern the organization of the hierarchy.

According to these guidelines, the object positioned at the lowest level of the spatial arrangement is designated as the root (parent) node, while subordinate nodes are assigned progressively toward the object located at the highest position. Each node was constrained to have only a single parent, ensuring a clear and unambiguous hierarchical structure.

This study represents a significant advancement in enhancing the robot’s reasoning ability based on a given task prompt. By incorporating a hierarchical tree structure, the model not only organized objects more effectively but also demonstrated a sophisticated understanding of how objects should be arranged according to specific tasks and goals (as illustrated in Figure~\ref{fig3a} and Figure~\ref{fig3b}).

In many \ac{VLMs}, spatial relationships between objects are not fully understood, which limits their reasoning capabilities. However, the hierarchical structure we implemented enabled the language model to reason more effectively and accomplish the given tasks. This approach significantly improved the robot's ability to adapt to task requirements while interacting with its environment in a structured and safe manner.

The methodology provides a framework for advancing spatial reasoning capabilities for robotic manipulation tasks. It provides a robust foundation for further research into improving the autonomy and problem-solving capabilities of robots, paving the way for more advanced systems that can adapt to complex and dynamic environments with greater autonomy.

\section{Conclusions}

We introduce a new approach to robotic manipulation that addresses current limitations. By developing a new dataset that includes object details along with manually defined spatial relationships, we provide a comprehensive framework for improving robotic interaction with objects in a task-oriented context. Using a hierarchical tree structure to represent these relationships enables more accurate and context-aware manipulations when integrated into a language model. By closing a critical gap in the literature on spatial reasoning in robotic manipulation, our method provides a scalable and adaptable solution for a wide range of applications. Furthermore, by making our dataset and code publicly available, we aim to encourage collaboration within the research community and enable further advances in the field.





\subsubsection{Acknowledgements.} Nurhan Bulus Guran is funded by Turkish Ministry of National Education, Republic of Türkiye.

\bibliographystyle{ieeetr}
\bibliography{references}

\begin{thebibliography}{10}

\bibitem{duan2024manipulate}
J.~Duan, W.~Yuan, W.~Pumacay, Y.~R. Wang, K.~Ehsani, D.~Fox, and R.~Krishna, ``Manipulate-anything: Automating real-world robots using vision-language models,'' {\em arXiv preprint arXiv:2406.18915}, 2024.

\bibitem{billard2019trends}
A.~Billard and D.~Kragic, ``Trends and challenges in robot manipulation,'' {\em Science}, vol.~364, no.~6446, p.~eaat8414, 2019.

\bibitem{zhu2023minigpt}
D.~Zhu, J.~Chen, X.~Shen, X.~Li, and M.~Elhoseiny, ``Minigpt-4: Enhancing vision-language understanding with advanced large language models,'' {\em arXiv preprint arXiv:2304.10592}, 2023.

\bibitem{liu2024visual}
H.~Liu, C.~Li, Q.~Wu, and Y.~J. Lee, ``Visual instruction tuning,'' {\em Advances in neural information processing systems}, vol.~36, 2024.

\bibitem{zheng2024survey}
Y.~Zheng, L.~Yao, Y.~Su, Y.~Zhang, Y.~Wang, S.~Zhao, Y.~Zhang, and L.-P. Chau, ``A survey of embodied learning for object-centric robotic manipulation,'' {\em arXiv preprint arXiv:2408.11537}, 2024.

\bibitem{Kruzliak_2024}
A.~Kruzliak, J.~Hartvich, S.~P. Patni, L.~Rustler, J.~K. Behrens, F.~J. Abu-Dakka, K.~Mikolajczyk, V.~Kyrki, and M.~Hoffmann, ``Interactive learning of physical object properties through robot manipulation and database of object measurements,'' in {\em 2024 IEEE/RSJ International Conference on Intelligent Robots and Systems (IROS)}, p.~7596–7603, IEEE, Oct. 2024.

\bibitem{gao2024physically}
J.~Gao, B.~Sarkar, F.~Xia, T.~Xiao, J.~Wu, B.~Ichter, A.~Majumdar, and D.~Sadigh, ``Physically grounded vision-language models for robotic manipulation,'' in {\em 2024 IEEE International Conference on Robotics and Automation (ICRA)}, pp.~12462--12469, IEEE, 2024.

\bibitem{chen2024spatialvlmendowingvisionlanguagemodels}
B.~Chen, Z.~Xu, S.~Kirmani, B.~Ichter, D.~Driess, P.~Florence, D.~Sadigh, L.~Guibas, and F.~Xia, ``Spatialvlm: Endowing vision-language models with spatial reasoning capabilities,'' 2024.

\bibitem{cheng2024spatialrgptgroundedspatialreasoning}
A.-C. Cheng, H.~Yin, Y.~Fu, Q.~Guo, R.~Yang, J.~Kautz, X.~Wang, and S.~Liu, ``Spatialrgpt: Grounded spatial reasoning in vision language models,'' 2024.

\bibitem{liao2024reasoningpathsreferenceobjects}
Y.-H. Liao, R.~Mahmood, S.~Fidler, and D.~Acuna, ``Reasoning paths with reference objects elicit quantitative spatial reasoning in large vision-language models,'' 2024.

\bibitem{driess2023palmeembodiedmultimodallanguage}
D.~Driess, F.~Xia, M.~S. Sajjadi, C.~Lynch, A.~Chowdhery, B.~Ichter, A.~Wahid, J.~Tompson, Q.~Vuong, T.~Yu, {\em et~al.}, ``Palm-e: An embodied multimodal language model,'' {\em arXiv preprint arXiv:2303.03378}, 2023.

\bibitem{brohan2023rt2visionlanguageactionmodelstransfer}
A.~Brohan, N.~Brown, J.~Carbajal, Y.~Chebotar, X.~Chen, K.~Choromanski, T.~Ding, D.~Driess, A.~Dubey, C.~Finn, {\em et~al.}, ``Rt-2: Vision-language-action models transfer web knowledge to robotic control,'' {\em arXiv preprint arXiv:2307.15818}, 2023.

\bibitem{wen2024objectcentricinstructionaugmentationrobotic}
J.~Wen, Y.~Zhu, M.~Zhu, J.~Li, Z.~Xu, Z.~Che, C.~Shen, Y.~Peng, D.~Liu, F.~Feng, and J.~Tang, ``Object-centric instruction augmentation for robotic manipulation,'' 2024.

\bibitem{jiang2023roboticperceptiontransparentobjects}
J.~Jiang, G.~Cao, J.~Deng, T.-T. Do, and S.~Luo, ``Robotic perception of transparent objects: A review,'' 2023.

\bibitem{blender2023}
{Blender Online Community}, ``Blender - a 3d modelling and rendering package.'' \url{https://www.blender.org}, 2023.

\bibitem{yolov8_ultralytics}
G.~Jocher, A.~Chaurasia, R.~Stoken, J.~Borovec, A.~NanoCode012, S.~Kwon, J.~Fang, J.~Fang, L.~Yu, T.~Laughing, A.~Hogan, I.~Imyhxy, A.~Suleiman, N.~Kharshiladze, F.~Ballesteros, L.~Bronk, , {\em et~al.}, ``Yolo by ultralytics.'' \url{https://github.com/ultralytics/ultralytics}, 2023.

\bibitem{openai2023gpt4}
OpenAI, ``Gpt-4,'' 2023.
\newblock Large language model, accessed September 24, 2024, \url{https://openai.com/gpt-4}.

\end{thebibliography}

\end{document}